\documentclass{article}

\usepackage{PRIMEarxiv}

\usepackage[dvipsnames]{xcolor}
\usepackage[utf8]{inputenc} 
\usepackage[T1]{fontenc}    
\usepackage{hyperref}       
\usepackage{url}            
\usepackage{booktabs}       
\usepackage{amsfonts}       
\usepackage{nicefrac}       
\usepackage{microtype}      
\usepackage{lipsum}
\usepackage{fancyhdr}       
\usepackage{graphicx}       
\graphicspath{{media/}}     

\usepackage{float}
\usepackage{multirow}
\usepackage{tabularx}
\usepackage{caption}
\usepackage{hyperref}

\usepackage{adjustbox}
\usepackage{makecell}
\usepackage{paralist}
\usepackage{relsize}

\title{CAD2DMD-SET: Synthetic Generation Tool of Digital Measurement Device CAD Model Datasets for fine-tuning Large Vision-Language Models
}

\author{
  João Valente \\
  Institute for Systems and Robotics \\
  University of Lisbon \\
  Lisbon, Portugal\\
  \texttt{joao.f.valente@tecnico.ulisboa.pt} \\
  \And
  Atabak Dehban \\
  Institute for Systems and Robotics \\
  University of Lisbon \\
  Lisbon, Portugal\\
  \texttt{adehban@isr.tecnico.ulisboa.pt} \\
  \And
  Rodrigo Ventura \\
  Institute for Systems and Robotics \\
  University of Lisbon \\
  Lisbon, Portugal\\
  \texttt{rodrigo.ventura@isr.tecnico.ulisboa.pt} \\
}

\begin{document}
\maketitle

\begin{abstract}

Recent advancements in Large Vision-Language Models~(LVLMs) have demonstrated impressive capabilities across various multimodal tasks.
They continue, however, to struggle with trivial scenarios such as reading values from Digital Measurement Devices~(DMDs), particularly in real-world conditions involving clutter, occlusions, extreme viewpoints, and motion blur; common in head-mounted cameras and Augmented Reality~(AR) applications.
Motivated by these limitations, this work introduces CAD2DMD-SET, a synthetic data generation tool designed to support visual question answering~(VQA) tasks involving DMDs.
By leveraging 3D CAD models, advanced rendering, and high-fidelity image composition, our tool produces diverse, VQA-labelled synthetic DMD datasets suitable for fine-tuning LVLMs.
Additionally, we present DMDBench, a curated validation set of 1,000 annotated real-world images designed to evaluate model performance under practical constraints.
Benchmarking three state-of-the-art LVLMs using Average Normalised Levenshtein Similarity~(ANLS) and further fine-tuning LoRA's of these models with CAD2DMD-SET's generated dataset yielded substantial improvements, with InternVL showcasing a score increase of 200\% without degrading on other tasks.
This demonstrates that the CAD2DMD-SET training dataset substantially improves the robustness and performance of LVLMs when operating under the previously stated challenging conditions.


The CAD2DMD-SET tool is expected to be released as open-source once the final version of this manuscript is prepared, allowing the community to add different measurement devices and generate their own datasets.


\end{abstract}

\section{Introduction}

\subsection{Motivation}\label{sec:motivation}

\begin{figure}[t]
\centering
\caption{DMDBench Challenging Examples for InternVL (a) and LLaVa (b). \textcolor{Green}{Green text} is correct answer, \textcolor{red}{red text} shows wrong answer.}
\label{tab:analysis}
\begin{tabular}{@{}p{0.48\textwidth}@{\hskip 0.04\textwidth}p{0.48\textwidth}@{}}

\begin{minipage}[t]{\linewidth}
\centering
\includegraphics[width=0.6\linewidth]{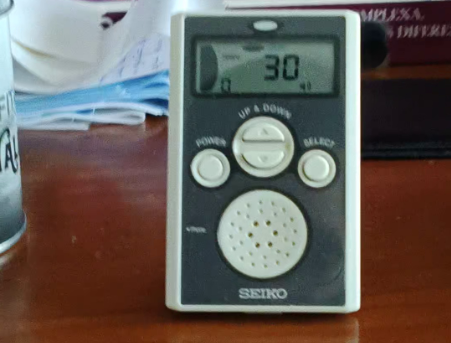}
\vspace{0.5em}

\smaller{\textbf{(a)Question:} What information is being conveyed by the digital display in this image?}
\end{minipage}
&
\begin{minipage}[t]{\linewidth}
\centering
\includegraphics[width=0.6\linewidth]{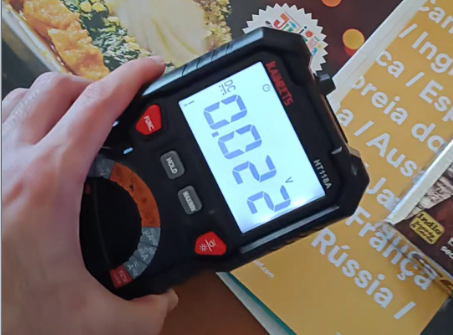}
\vspace{0.5em}

\smaller{\textbf{(b)Question:} Identify the device and summarise what it is displaying.}
\end{minipage}
\\
\midrule

\multicolumn{2}{l}{\smaller \textbf{Ground-Truth Answer:}} \\
\smaller The digital display conveys a TEMPO reading of 30 BPM. & \smaller The image shows a digital multimeter with a reading of 0.022 V. \\
\midrule

\multicolumn{2}{l}{\smaller \textbf{Non-fine-tuned Model Answer:}} \\ 
\smaller The digital display shows the number '\textcolor{Green}{30}', which could indicate a \textcolor{red}{timer setting}, a \textcolor{red}{countdown}, or a measurement reading, depending on the context of the device's use. & \smaller The device is a \textcolor{red}{digital clock}, and it is displaying the time as \textcolor{red}{00:22}. \\
\midrule

\multicolumn{2}{l}{\smaller \textbf{Fine-tuned Model Answer:}} \\
\smaller \textcolor{Green}{The digital display conveys a TEMPO reading of 30 BPM.} & \smaller The device is a \textcolor{Green}{digital multimeter}, and it is displaying \textcolor{Green}{0.022} \textcolor{red}{A}. \\
\end{tabular}
\end{figure}

Despite the impressive progress of the current state-of-the-art large vision-language models~(LVLMs) and their broad applicability in various domains, their implementation in real-world applications can present unforeseen challenges~\cite{jiang2024effectiveness}.
One example is reading from measurement devices, which is effortlessly performed by humans.
The examples below, shown in Fig.~\ref{tab:analysis}, exemplifies these difficulties.


One contributing factor is the scarcity of open-source datasets expressly designed for measurement-device reading.
Most large image–text corpora scraped from the internet are product-grade, displaying high-quality photos with minimal background clutter and frequently showing the same reading, often zero.
A related problem was studied by~\cite{vo2025vision}.
These conditions contrast with real-world scenes, where motion blur, visual clutter, and diverse~(non-zero) readings prevail.
When DMD readings are captured by head-mounted cameras~(\textit{e.g.} in AR applications), oblique viewpoints, and continuous motion blur intensify these difficulties~\cite{my_paper}.
 



\subsection{Objectives and Contributions}\label{sec:topic_overview}

Enabling LVLMs to read measurements from DMDs poses distinct challenges due to the diversity of display types~(LED, LCD, etc.), measurement units, and frequent use of symbols and abbreviations.
The absence of a dedicated dataset has also hindered progress in this domain.
This research aims to bridge that gap, offering a foundation for advancing OCR and VQA methods tailored to digital measurement interpretation.

\noindent To summarise, this work has four main contributions:
\begin{itemize}
    \item We provide CAD2DMD-SET, a flexible synthetic data generation tool targeted at creating a large dataset of DMD;
    \item We provide a labelled real validation set~(DMDBench) of multiple DMDs and benchmark various state of the art open-weight LVLMs;
    \item We created a large synthetic dataset to fine-tune multiple LVLMs using CAD2DMD-SET;
    \item We experimentally demonstrate the applicability of the tool in improving the performance of LVLMs in reading DMDs from unconstraint view points and analyse the sources of improvements.
\end{itemize}



\section{Literature Review}

\subsection{Photogrammetry-based 3D Reconstruction}

\textbf{Photogrammetry-based 3D reconstruction} is a technique used to create 3D models of real-world objects or environments from a series of 2D photographs.
It works by analyzing multiple images of the same object or scene taken from different angles and then using algorithms to triangulate the 3D coordinates of points in space.
Several applications allow for such reconstruction and our two main options consisted of \textbf{KIRI Engine}, from KIRI Innovations and \textbf{Object Capture}, from Apple.

KIRI Engine, designed for ease of use across mobile and desktop platforms, offers an accessible and cost-effective alternative to traditional scanning methods by relying solely on image data, supporting high-resolution texture mapping and mesh generation.

Object Capture, which is part of the RealityKit framework on macOS, relies on structure-from-motion~(SfM) and multi-view stereo~(MVS) techniques to estimate camera poses and reconstruct dense point clouds, which are subsequently meshed and textured to produce high-fidelity 3D assets.
Compared to traditional photogrammetry pipelines, which often require specialised software and substantial manual tuning, Object Capture emphasises ease of use and automation.
Unlike KIRI Engine, the framework is optimised for Apple’s hardware ecosystem only, making use of GPU-accelerated computation and native integration with macOS APIs.
Furthermore, it achieves results competitive with established photogrammetry tools.
Although Object Capture uses SfM techniques for 3D asset reconstructions, some methods use Guassian Splatting \cite{guassian_paper}.
We decided not to pursue such methods, however, because these 3D models do not have a mesh or a surface to unwrap, making UV mapping~(a critical step in CAD2DMD-SET) unfeasible. 

While KIRI Engine is compatible with multiple platforms, Object Capture offers a \textbf{free}, \textbf{open-source} alternative that produced superior overall model quality.
Its output featured cleaner, more organised meshes, which facilitated easier and quicker manual refinement of our 3D models, when needed.
As such, our final choice for a photogrammetry-based 3D reconstruction tool was Object Capture.

\subsection{3D Modeling and Rendering}

Today, a broad selection of free \textbf{3D modeling and rendering} software is readily accessible.
Fundamentally, modeling tools enable the accurate creation of shapes and structures, while rendering engines replicate the effects of lighting and materials to generate realistic or stylised visuals.
These tools are extensively employed across various industries, and technological advancements have significantly enhanced their capabilities and availability.
Two main options were initially considered, \textbf{Unity's  Perception Package}~\cite{Unity_Perception} and \textbf{Blender}.
However, Unity’s Perception package was officially discontinued and is now community-maintained only, which makes it a less reliable choice. 

\textbf{Blender}, on the other hand, is a comprehensive, \textbf{open-source} software suite for 3D modeling, animation, rendering, and simulation.
It was originally developed as an in-house application by NeoGeo and later released to the public under the GNU General Public License.
Overall, Blender has evolved into a robust and widely adopted platform within both academic and industrial domains.
Its open-source nature not only ensures cost-effectiveness but also allows for extensive customisation and community-driven enhancements, making it particularly suitable for research and development purposes.
Furthermore, Blender supports a wide range of functionalities including mesh modeling, UV mapping, texturing, real-time rendering and more.
The integration of python scripting further extends its utility, enabling automated workflows and seamless integration with external tools and data pipelines.

With this in mind, Blender was used for the development of the CAD2DMD-SET automation tool, via \textbf{python scripting}.
From the available render engines, Eevee Next, a rasterisation-based engine, was chosen primarily due to its rendering speed, while still providing sufficiently realistic renders for the generation of synthetic data.
For that reason, Eeeve Next provided the optimal balance of quality and performance in tasks that are time-consuming, such as large synthetic data generation.

\subsection{Image Composition}

In computer vision, \textbf{image composition} is the process of integrating a foreground object into a background image to create a seamless and visually realistic composite.
This involves several challenges, which range from \textbf{appearance inconsistency}, such as mismatches in colour, lighting, and texture to \textbf{geometric inconsistency}, which includes misalignment in perspective and/or scale; and \textbf{semantic inconsistency}, where the inserted object may not logically fit within the context of the scene. 

The library of Image Composition, \textbf{libcom}~\cite{Libcom} is a dedicated toolbox developed to support and advance research in this area.
It provides a structured framework that encompasses a diverse set of image composition tasks, including image blending for smooth boundary transitions, image harmonisation to correct stylistic and photometric differences, shadow generation to enhance realism through consistent lighting cues, and object placement to ensure spatial and contextual plausibility.
It also incorporates generative composition techniques, which leverage data-driven models to synthesise realistic integrations, and quality evaluation modules that assess the visual fidelity of the composite output.

Two models are available in libcom, consisting of \textbf{ControlCom-Image-Composition} ~\cite{Control-Com}, a controllable image composition method that unifies four tasks in one diffusion model~(image blending, image harmonisation, view synthesis, and generative composition) and \textbf{FOPA}~(Fast Object Placement Assessment)~\cite{FOPA}, a discriminative method which can predict the rationality scores for all locations/scales given a background-foreground pair, thus determining the most contextually appropriate placements.
FOPA uses a dual-encoder architecture with dynamic filters to integrate foreground and background features.
It enhances performance by transferring knowledge from ~\cite{OPA} through weight initialization and feature alignment.


By offering a cohesive set of tools under a unified interface, libcom facilitates both the practical implementation and experimental evaluation of image composition techniques.
We chose libcom’s FOPA model as it delivers the speed of discriminative methods without the measurement-distorting risks of the generative ControlCom-Image-Composition alternative.


\subsection{LVLM Benchmarks}

While evaluating large multimodal models is crucial, performing comprehensive assessments across numerous benchmarks has proven to be highly burdensome.
This process often involves aggregating data from a wide array of repositories and navigating potential compatibility issues across different environments.

For these reasons, we utilised \textbf{VLMEvalKit}~\cite{VLMEvalkit} to perform the benchmarks of our LVLMs of choice.
VLMEvalKit is an open-source toolkit for evaluating large multi-modality models based on PyTorch that provides a unified framework capable of automating dataset handling, model interfacing, and result aggregation.
These enable efficient, \textbf{reproducible}, and extensible evaluation workflows, thus facilitating our comprehensive benchmark of LVLMs with minimal overhead.

\section{CAD2DMD-SET Data Generation Tool}

\begin{figure*}[tbp]
  \begin{center}
    \includegraphics[width=\linewidth]{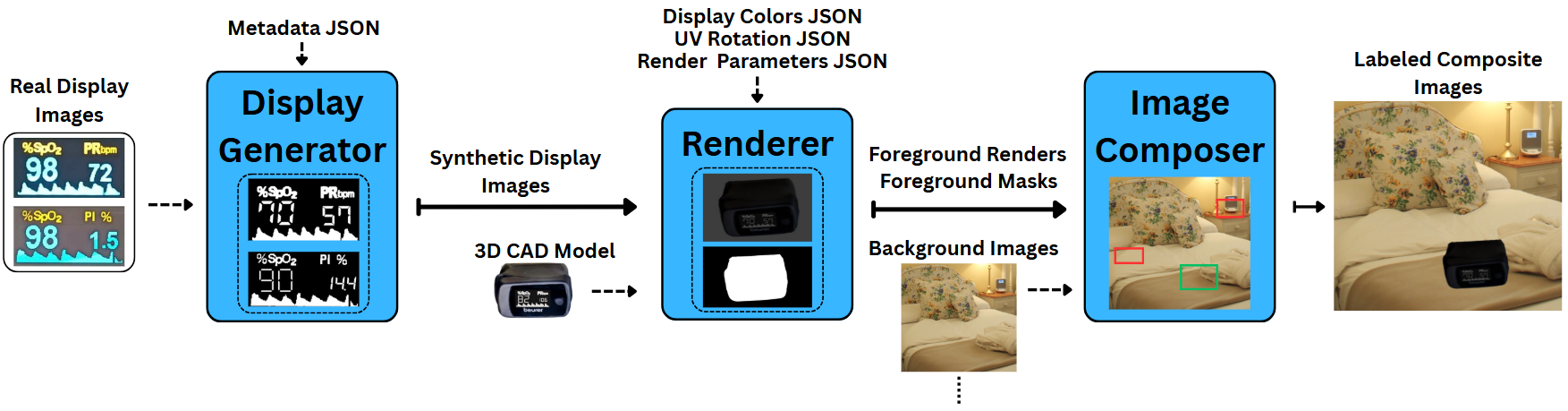}  
  \end{center}
  \caption{CAD2DMD-SET Pipeline - Pulse Oximeter Example}
  \label{fig:tool_pipeline}
\end{figure*}

This section provides a detailed overview of the individual modules that together make up the pipeline of our data generation tool, \textbf{CAD2DMD-SET}.
Each module is categorised by its main functionality, which addresses specific challenges associated with DMD OCR and VQA, as addressed in Section~\ref{sec:topic_overview}. 
CAD2DMD-SET's pipeline is depicted in Fig.~\ref{fig:tool_pipeline}.


The first component, the \textbf{Display Generator}, generates multiple display images, using measurement values sourced from dictionaries tailored to the specific DMD.
It also includes the generation of these dictionaries, which encapsulate the range of possible device readings.


The second component, the \textbf{Renderer}, utilises Blender’s Python API in conjunction with the 3D CAD models of the devices to produce base renderings.
These renderings include the default world background from Blender and incorporate variability in lighting, model orientation, and camera distance to enhance visual diversity.
Additionally, this module performs UV mapping to project synthetic display images onto the corresponding device surfaces and generates the associated VQA labels.
The resulting device renderings, along with their corresponding segmentation masks, also generated in the Renderer component, are subsequently used as foreground objects in the next module of the pipeline.

The final component, the \textbf{Image Composer}, employs the libcom library to create high-fidelity composite images by placing rendered CAD models within realistic indoor background scenes.
It leverages FOPA to determine the most contextually appropriate placement given a foreground and a background image and produces composite images by pasting the scaled renders onto these locations with computed realism.

\subsection{Display Generator}
The \textbf{Display Generator} module produces the various display images associated with each device.
Its operation comprises two primary stages: the generation of \textbf{dictionaries}, and the subsequent creation of the \textbf{display images}.

\begin{figure}[t]
\begin{minipage}[c]{0.5\textwidth}
    \centering
    \includegraphics[width = 6.5 cm]{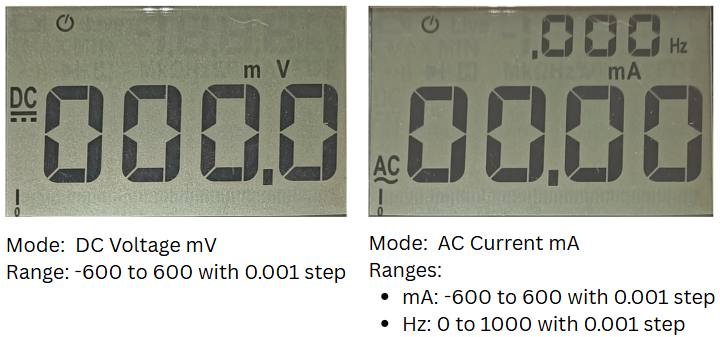}
    \caption{Real Multimeter Display.}
    \label{fig:real_display}
\end{minipage}
\hspace{0.3em}
\begin{minipage}[c]{0.45\textwidth}
    \centering
    \includegraphics[width = 7.5 cm]{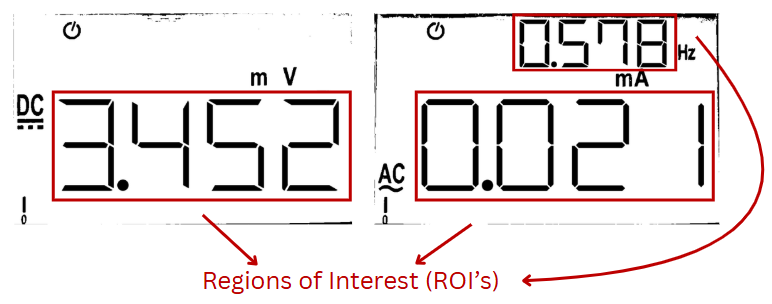}
    \caption{Synthetic Multimeter Display.}
    \label{fig:synthetic_display}
\end{minipage}
\end{figure}

First, the module has functionalities to automatically generate dictionaries in text file format, similar to the ones used in EasyOCR~\cite{EasyOCRGit}, which contain the range of measurements of a certain unit that can be placed on a DMD display.
DMDs may operate in multiple \textbf{modes}, each associated with distinct \textbf{measurement ranges}.
The mode of a device determines the set of measurable units available in that configuration, thereby influencing both the visual layout of the display and the range of values it can represent.
An illustrative example is provided in Fig.~\ref{fig:real_display}.


As a dictionary is linked to a specific measurement range, it can be shared between devices using the same range.
If new devices use different ranges, new dictionaries must be created.
In addition to our automatic functions, these dictionary files can also be generated externally and imported into CAD2DMD-SET as text files.

With the necessary dictionaries defined, the second stage of this module involves using the specified ranges to generate the display images with varying measurements, requiring one real image for \emph{each mode} of the DMDs included in CAD2DMD-SET.
Each image should contain only the display screen of the corresponding display, as seen in Fig.~\ref{fig:real_display}.

For each real display image and its corresponding device mode, a \texttt{JSON} file is associated which contains the metadata.
This metadata includes 
\begin{inparaenum}
    \item the image filename,
    \item the number of regions of interest~(which reflects how many simultaneous measurements are displayed in that specific mode),
    \item the device mode itself,
    \item and detailed label information.
\end{inparaenum}
The labels specify the types of measurements being shown~(\textit{e.g.} \emph{Perfusion Index} or \emph{Pulse Rate} in the case of the pulse oximeter) along with their respective units.
Notably, this label information is used exclusively in the subsequent module of CAD2DMD-SET for generating the VQA labels associated with the foreground object renderings.

Afterwards, we process each image of a DMD display to isolate its foreground from the background.
The image is first converted to greyscale to simplify the data processing and then the Otsu thresholding~\cite{otsu1979threshold} for binarization is applied.
This approach is particularly useful as it eliminates the need for manual threshold selection and enhances reproducibility across images.
The corresponding mode of the image is specified in the provided metadata \texttt{JSON} file.


The process of updating the digital display image consists of embedding new measurement values into predefined \textbf{regions of interest~(ROIs)} and saving the modified image. 
Some examples can be visualized in Fig.~\ref{fig:synthetic_display}. 
Metadata stored in a \texttt{JSON} file specifies the number of required ROIs based on the image mode. 
If ROI coordinates for the mode already exist, they are used directly, otherwise, the user is prompted to define them interactively. 
These coordinates are then stored in the \texttt{JSON} file for subsequent reuse.

Once the ROI coordinates are available, a binary threshold resets pixel values to match the background, clearing each region for clean text insertion and avoiding visual artifacts.
Next, a measurement value is randomly selected from the corresponding dictionary for each ROI.
These values are rendered onto the image by randomly selecting one of 24 available fonts from the DSEG family~\cite{DSEG}, which mimics the segmented style commonly found in electronic displays.
Finally, the measurement is rendered using the foreground colour, and the resulting synthetic display image is saved to a designated output directory.




\subsection{Renderer}

The primary function of the \textbf{Renderer} module is to programmatically stitch the display images and generate renderings of the device’s 3D CAD models in Blender, ensuring these include variability in lighting conditions, model orientation, and camera distance.
This structured and automated generation is possible via Blender Python API.


Prior to rendering, CAD models must be \emph{preprocessed} by ensuring the display comprises a single rectangular face, with possible manual modifications to the mesh being necessary to achieve such requirement. This face must be later selected in Blender, using a custom-developed operator ``ExportSelectedFace'' to export its index to a \texttt{JSON} file, which is later used for accurate UV mapping.
Additionally, new devices require their name and initial display rotation to be specified in a designated \texttt{JSON} file.


Following the preprocessing stage, the rendering parameter ranges (distance to camera, rotation around access, etc.) are specified in a separate configuration file.
This file specifies the allowable ranges and selectable options for key parameters that control the rendering environment, including, for example, the camera's relative distance to the object and the object's rotation about the x-axis, which were set between 8 to 10 meters and from -30$^\circ$ to 30$^\circ$, respectively.
By externalising these values into a structured configuration file, the system ensures reproducibility, flexibility, and ease of experimentation without requiring modifications to the core rendering code.


Given the configuration file, the rendering environment is configured to predefined specifications, including image resolution and file format and the 3D model of a device is programmatically loaded.
Then, we enter the \textbf{UV mapping} phase, where a texture representing a synthetic digital display, selected randomly from the dataset of display images~(pre-generated by the Display Generator), is applied to a specific face of the 3D model, identified beforehand.
The texture is then mapped onto the model using UV coordinates that are adjusted based on a rotation parameter, also from the preprocessing phase, to ensure proper alignment.
A separate \texttt{JSON} file defines a palette of ``near-grey'' colors, from which a background color is randomly chosen to introduce subtle chromatic variation and enhance visual diversity while maintaining the realistic appearance of DMDs.


Subsequently, the \textbf{camera} is \textbf{placed} considering both variability and full visibility of the 3D object within the rendered frame.
The camera is first orientated to look directly at the object by computing a vector from the object’s centre to a predefined offset location, ensuring that the optical axis of the camera aligns with this vector.
The camera is then placed at a distance from the object, proportional to its maximum bounding dimension, with the exact value randomly sampled within the predefined interval.
This proportional sampling ensures consistent framing across devices of varying sizes.
The current implementation also introduces variability in focal length, randomly selecting values from the specified range.

Following camera placement, the \textbf{object} undergoes a \textbf{stochastic rotation}.
Its orientation is initially reset to a neutral pose, after which a random rotation is applied along one of the principal axes~(X, Y, or Z).
The angle of rotation for the selected axis is drawn from predefined bounds that are configurable via the \texttt{JSON} external parameter file.

Before rendering proceeds, the system validates that the object is fully within the camera’s field of view.
If not, camera placement and object rotation are resampled until this condition is met, ensuring renders are free from occlusions or framing errors, necessary for use as foreground elements in the CAD2DMD-SET pipeline.


Once the camera and object are correctly configured, a \textbf{point light source} is \textbf{positioned} to illuminate the object from a random offset.
This offset is defined in terms of relative distances along the X, Y, and Z axes, scaled by the object’s largest dimension to maintain consistent lighting geometry across different model sizes.
The light’s colour is selected from a constrained random space, favouring near-white hues to simulate ambient indoor lighting.
Intensity~(energy), falloff softness, and light radius are also randomised within the specified limits.
These parameters modulate the harshness and spread of shadows and screen glare-like effects, adding realistic variability to the illumination conditions.

Altogether, this pipeline generates a diverse array of physically plausible foreground DMDs, each with randomised but controlled parameters for camera position, object rotation, and lighting.
Once the scene is fully configured, rendering is executed with multiple output passes.
The rendering engine captures not only the final \textbf{RGB image}, but also intrinsic passes such as the \textbf{depth map}, and optionally albedo, shading, and normal maps.
The depth pass is normalised using a mapping function in Blender’s compositor, scaled between predefined depth bounds, and exported separately to facilitate mask generation.


Post-processing starts after the rendering phase, with \textbf{motion blur} being applied to foreground images to simulate dynamic capture conditions, such as device or camera movement.
Motion blur is then applied by convolving the image with a randomly oriented and sized kernel to introduce diverse motion artefacts. This is followed by the second post-processing step: the generation of binary segmentation masks from the depth maps.
A \texttt{CSV} file records all render parameters to ensure scene reproducibility, and this metadata is used to automatically generate question-and-answer labels, enabling efficient creation of diverse annotations without requiring additional, time-costly renders.


\subsection{Image Composer}

The \textbf{Image Composer}, the last module of CAD2DMD-SET, is responsible for generating the final composite images that make up the synthetic dataset of DMDs.
Three prerequisites are needed for its functioning: \textbf{foreground object images}, their corresponding \textbf{masks}, and a set of \textbf{background images}, on which the scaled foreground objects will be placed. 

In selecting background images, two indoor scene images datasets were initially considered: the Indoor Scene Recognition MIT dataset~\cite{indoor_mit} and the ADE20k~\cite{ade20k}.
However, due to the relatively low resolution of most images in MIT's dataset~(256×256 pixels), a curated subset of approximately 2,900 high-resolution indoor scene images from ADE20k was ultimately used.
These images exceed 600x800 pixels, offering a higher visual quality.
Ensuring high resolution for background images is crucial, as it not only determines the resolution of the final composite image but also directly influences the visual fidelity of the foreground object integrated into the scene.

When all prerequisites are present, the Image Composer module takes advantage of two main functionalities from the libcom library: the FOPA model and its naive composition api which generates composite images through copy-and-paste.



Unlike SOPA, FOPA's architecture enables rapid inference across all spatial locations simultaneously.
This makes it especially valuable for our data generation tool, where fast and consistent evaluation of numerous placements is essential.
The inference time of FOPA depends on the resolution of the background image, with  higher-resolution backgrounds leading to longer inference times.
To balance visual quality and inference speed, the high-resolution background images from ADE20K are downscaled—setting the longest side to 256 pixels while preserving aspect ratio—so that FOPA can process them more efficiently without sacrificing foreground detail.
The model then predicts placements on the downscaled background, and the results are then rescaled to fit the original high-resolution image, combining fast inference with high-resolution output.

The resulting scaled bounding box then undergoes two adjustments:~\textbf{aspect ratio correction} and \textbf{scale validation}.
FOPA's bounding box predictions often distort foreground objects' aspect ratio, producing unrealistic results.
We preserve the original aspect ratio by computing it from the binary mask and adjusting the box width while keeping height fixed, centered at the original location. 
The box is scaled up if it covers less than 10\% of the background area.
We track all triplets to prevent duplicates.
While these resizing and reshaping may slightly reduce realism, they are essential for maintaining foreground readability.


The final, adjusted bounding box is then passed to a custom libcom function, which generates the composite image by copying and pasting the foreground object at the specified location and scale defined by the bounding box.



Initial experiments with image harmonisation~\cite{hu2023image} showed it reduced the clarity of digital displays by blurring or distorting the digits, compromising their accuracy and usability. However, as more advanced models are continuously being proposed, the tool includes this feature as an optional setting that can be enabled by the user.

To further accelerate the composite data generation process, the Image Composer supports multiprocessing, enabling multiple workers and GPUs to operate in parallel, thus achieving an average of 1.4s per generated composite image.

Throughout the data generation process, a \texttt{CSV} file is also maintained to log essential information for each composite image.
This includes filenames for the composite image, foreground, foreground mask, and background, along with the bounding box coordinates corresponding to the foreground object placement.

Finally, VQA pairs are generated for each composite image by referencing this \texttt{CSV} file.
For each entry, the associated foreground object is retrieved, and its pregenerated labels are used to construct the VQA annotations, associated to the corresponding composite image.

\section{Experiments and Results}

This section outlines the DMDs utilised in generating the fully synthetic training dataset.
It also details the development and annotation of the corresponding validation set, \textbf{DMDBench}, composed of only real images, which was used to benchmark \textbf{three} different LVLMs: Pixtral-12B~\cite{Pixtral12B}, LLaVA-1.5-13B~\cite{Llava}~(henceforth LLaVA), and InternVL2.5-26B~\cite{InternVL}~(henceforth Intern).
Notably, LLaVa and InternVL were evaluated both before and after fine-tuning.
All LVLM's were fine-tuned following the \textbf{two-stage tuning paradigm}~\cite{lvlm_training_paradigm}, in which a second tuning phase is added~(LoRA~\cite{LoRA} modules, in our case) to refine performance with a manageable number of parameters. Each LoRA was trained with a mixture of the model's standard instruction tuning data and our synthetic dataset.

\begin{figure}[H]
    \centering
    \includegraphics[width=0.9\linewidth]{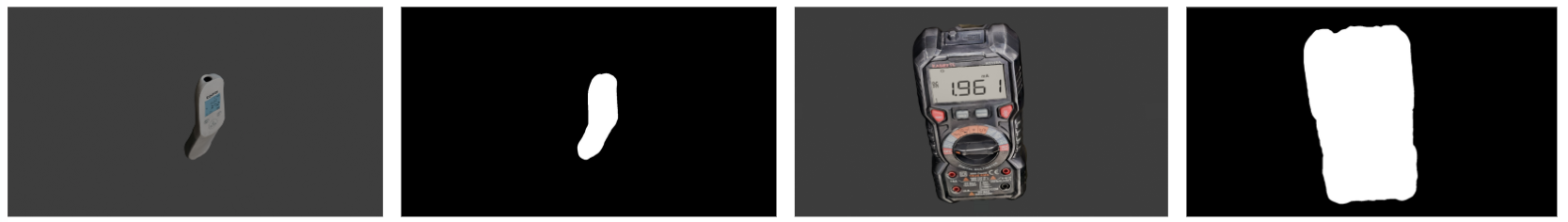}
    \caption{Foreground Object Renders and Masks~(Thermometer and Multimeter).}
    \label{fig:foreground_and_masks}
\end{figure}

\subsection{Training Dataset}

To generate the synthetic training dataset, \textbf{six} DMDs were selected, including a power supply and a multimeter, commonly used in industrial settings, a metronome from the music industry, and three medical instruments: a thermometer, a pulse oximeter, and a blood pressure monitor.
These devices were selected based on their accessibility and diverse set of applications, thus having varying display styles to improve the generalization capabilities of the LVLMs.
Their 3D CAD models were reconstructed using photogrammetry through Object Capture, with approximately 200 images captured from multiple distances and angles for each device.
For visualisation, the 3D CAD model renderings of two devices and their masks are illustrated in Fig.~\ref{fig:foreground_and_masks}.

\begin{figure}[t]
    \centering
    \includegraphics[width=0.8\linewidth]{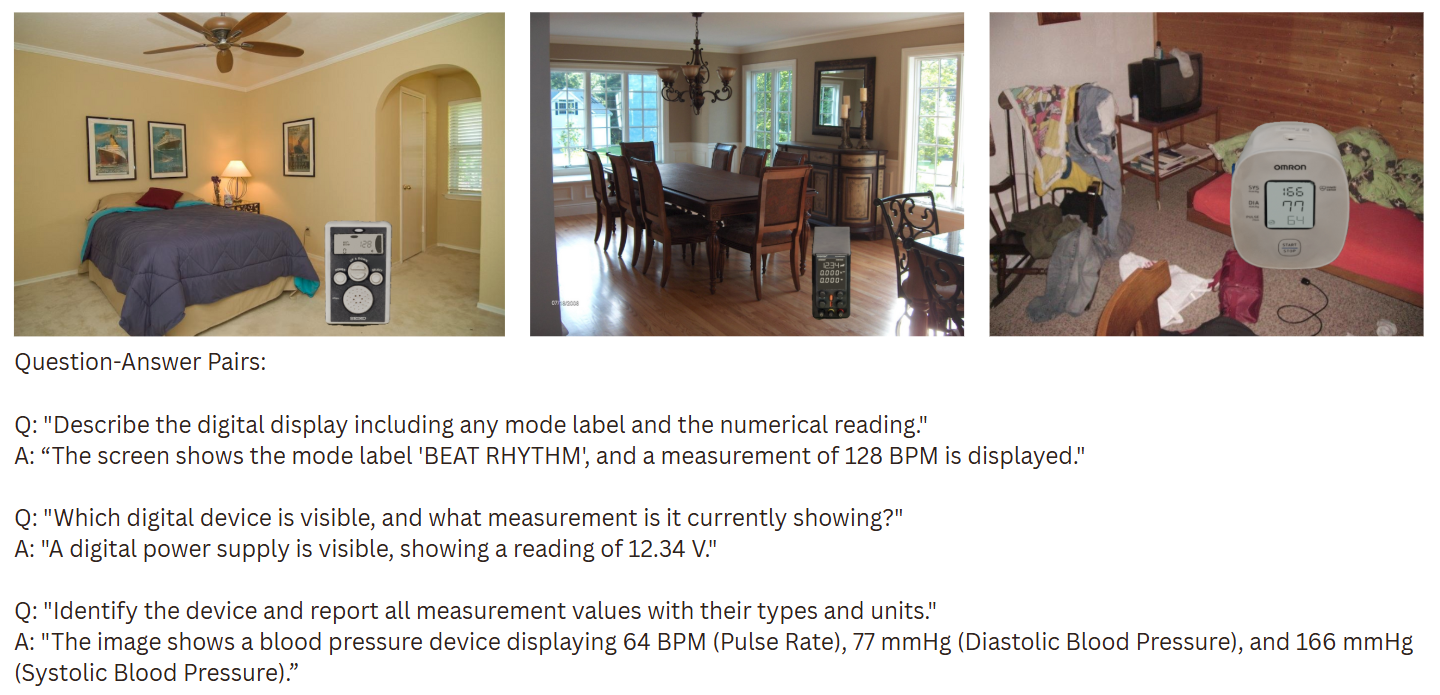}
    \caption{Examples of the Metronome, Power Supply and Blood Pressure Device.}
    \label{fig:composite}
\end{figure}

Finally, we used the CAD2DMD-SET tool to generate the \textbf{training dataset}, composed of \textbf{100k} composite images, with VQA labels to be used for LVLMs fine-tuning.
Examples of final labelled composite images can be seen in Fig.~\ref{fig:composite}.

\subsection{Validation Dataset - DMDBench}

\textbf{DMDBench} was collected using \textbf{1,000} manually annotated real images of selected DMDs with information on the type of device, its operating mode, and visible measurements.



\begin{figure}[H]
    \centering
    \includegraphics[width=0.9\linewidth]{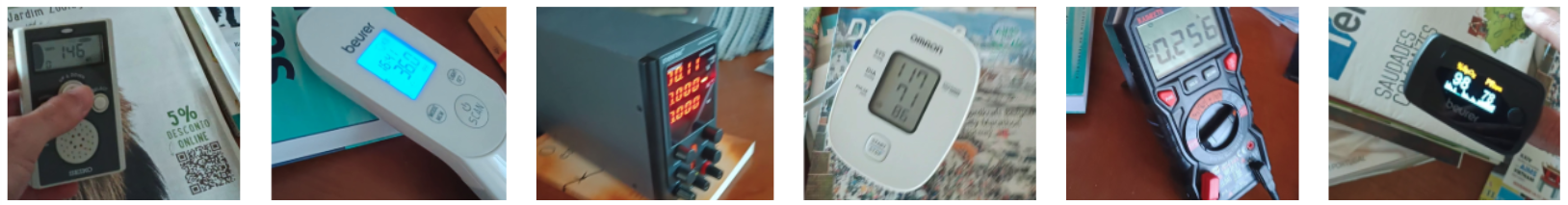}
    \caption{DMDBench Image Examples}
    \label{fig:dmdbench}
\end{figure}

Using the labeling tool developed within the CAD2DMD-SET system, we generated question-answer pairs for each image, mirroring the approach used to produce VQA-style labels for synthetic training images.
The images in DMDBench are intended to emulate realistic and imperfect conditions, incorporating variations in scale, orientation, lighting, background clutter, and motion blur caused by user movements, characteristics commonly encountered in head-mounted AR devices. DMDBench examples are illustrated in Fig.~\ref{fig:dmdbench}.

\subsection{Benchmarks and Fine-Tuning}




For consistency and comparability, all benchmarks were conducted on an NVIDIA A100 GPU.
Given that DMDBench is a VQA dataset, the \emph{Average Normalised Levenshtein Similarity~(ANLS)}~\cite{anls_metric} was adopted as the main evaluation metric.
ANLS quantifies how closely a model's predicted answer matches the ground truth, accounting for partial correctness. 
It operates by computing the Levenshtein distance, which is then normalised and converted into a similarity score ranging from 0 to 1, where 1 denotes an exact match.
We decided on ANLS as the evaluation metric, since it tolerates minor variations in wording or phrasing that do not significantly alter the semantic content of an answer, making it particularly well-suited for VQA tasks~(see Fig.~\ref{tab:analysis}).
This makes it a more robust and informative metric than strict accuracy when evaluating open-ended responses. 

To assess the relative difficulty that DMDBench images present to LVLMs, the same models were also evaluated on the TextVQA dataset~\cite{TextVQA}.
This a benchmark specifically designed to test the ability of models to read and reason about text within images.
By comparing model performance across both DMDBench and TextVQA, it becomes possible to contextualise the challenges posed by DMDBench, particularly in terms of the model's capacity to integrate visual and textual information under real-world conditions.
These benchmarks, executed by using VLMEvalKit, can be viewed in Table~\ref{tab:full_benchmark}.






\begin{table}[H]

\caption{LVLM Benchmarks}
\label{tab:full_benchmark}

\begin{adjustbox}{width=1\textwidth}
\begin{tabular}{cc|c|cccc|}
\cline{3-7}
                                                           &                                   & \textbf{TextVQA}   & \multicolumn{4}{c|}{\textbf{DMDBench}}                                                                                                                                            \\ \hline
\multicolumn{1}{|c|}{\textbf{Model Version}}               & \makecell{\textbf{Training} \\  \textbf{Data Usage (\%)}} & \textbf{ANLS (\%)} & \multicolumn{1}{c|}{\textbf{ANLS (\%)}} & \multicolumn{1}{c|}{\makecell{\textbf{Word-level} \\ \textbf{Accuracy}}} & \multicolumn{1}{c|}{\makecell{\textbf{Unit} \\ \textbf{Accuracy}}} & {\makecell{\textbf{Numeric} \\ \textbf{Accuracy}}} \\ \hline
\multicolumn{1}{|c|}{\textbf{Pixtral-12B}}                 & -                                 & 41.64              & \multicolumn{1}{c|}{32.21}              & \multicolumn{1}{c|}{-}                            & \multicolumn{1}{c|}{-}                            & -                               \\ \hline
\multicolumn{1}{|c|}{\textbf{LLaVA-v1.5-13B}}              & -                                 & 17.11              & \multicolumn{1}{c|}{25.06}              & \multicolumn{1}{c|}{6.9}                          & \multicolumn{1}{c|}{3.7}                          & 10.6                            \\ \hline
\multicolumn{1}{|c|}{\textbf{InternVL2.5-26B}}             & -                                 & 76.96              & \multicolumn{1}{c|}{32.92}              & \multicolumn{1}{c|}{65}                           & \multicolumn{1}{c|}{86.2}                         & 50.8                            \\ \hline
\multicolumn{1}{|c|}{\textbf{LLaVA-v1.5-13B (Fine-tuned)}}  & 10                                & 17.28              & \multicolumn{1}{c|}{71.21}              & \multicolumn{1}{c|}{34.1}                         & \multicolumn{1}{c|}{55.3}                         & 17.1                            \\ \hline
\multicolumn{1}{|c|}{\textbf{LLaVA-v1.5-13B (Fine-tuned)}}  & 50                                & 14.96              & \multicolumn{1}{c|}{53.55}              & \multicolumn{1}{c|}{30.2}                         & \multicolumn{1}{c|}{36.8}                         & 26.7                            \\ \hline
\multicolumn{1}{|c|}{\textbf{InternVL2.5-26B (Fine-tuned)}} & 10                                & \textbf{77.03}     & \multicolumn{1}{c|}{95.16}              & \multicolumn{1}{c|}{69.4}                         & \multicolumn{1}{c|}{84.5}                         & 61.4                            \\ \hline
\multicolumn{1}{|c|}{\textbf{InternVL2.5-26B (Fine-tuned)}} & 50                                & 76.76              & \multicolumn{1}{c|}{\textbf{96.04}}     & \multicolumn{1}{c|}{\textbf{75.1}}                & \multicolumn{1}{c|}{\textbf{89.2}}                & \textbf{68.6}                   \\ \hline
\end{tabular}
    
\end{adjustbox}

\end{table}

Considering the non-fine-tuned model versions, \textbf{DMDBench} appears to be the \textbf{more challenging} benchmark, as evidenced by the drop in performance for most models compared to TextVQA, with the exception of LLaVa.
For instance, InternVL sees a decrease from 76.96\% on TextVQA to just 32.92\% on DMDBench, suggesting that the latter likely includes more complex or ambiguous text-based reasoning tasks.

For both LLaVa and InternVL, a LoRA was fine-tuned and its weights merged with the original model in order to obtain a final fine-tuned version. Two LoRA's were fine-tuned per model, with 10\% and 50\%, respectively, of a mixture of our synthetic dataset and the standard instruction tuning data, which differs for each model.
In LLaVa's case, it includes a mix of images from COCO~\cite{COCO}, GQA~\cite{GQA}, OCR-VQA~\cite{OCR-VQA}, TextVQA~\cite{TextVQA} and VisualGenome~\cite{VG} datasets. For InternVL, it encompasses LLaVA's instruction tuning data and also AI2D~\cite{ai2d}, ChartQA~\cite{chartqa}, DocVQA~\cite{docvqa}, DVQA~\cite{dvqa}, LLaVA-Pretrain, SAM~\cite{sam}, SynthDoG-EN~\cite{donut}, WebData and GeoQA+~\cite{geoqa} datasets.
The decision to use only a fraction of the available data was motivated by initial experiments with fine-tuning LLaVa using 100\% of the training set, which led to representation collapse, a phenomenon where the model’s output distribution becomes degenerate, severely limiting its expressive capacity.
This is inline with the training stability observed during fine-tuning large language models~\cite{stability}.



Comparing the fine-tuned version of both models with their non-fine-tuned counterparts, we can verify that the addition of our synthetic data to the mixture used during the instruction tuning phase drastically \textbf{increases} the ANLS score, on DMDBench.
For instance, when trained with 50\% of its training data mixture, InternVL sees an ANLS score increase from 32.92\% to 96.04\%.
Analysing the results, we attribute the improvements in ANLS score to the following two corrections in the fine-tuned version: incorrect identification of the DMD, based on readable measurements, in cases where the latter are difficult to read due to presence of motion blur or poor lighting conditions; and wrong measurement reading, due to scene text environmental conditions.

The ANLS score improvement /indicates the model learned proper sentence structures from the synthetic dataset Fig.~\ref{tab:analysis}. 
Furthermore, stable TextVQA performance shows our fine-tuning avoids catastrophic forgetting on external datasets..

The comparison of two fine-tuned InternVL models on DMD-Bench shows that a 40\% increase in training data yields only a minor ANLS improvement, suggesting that using just 10\% of the data may be more efficient in resource-limited settings, however, this trend does not hold for all models, as the fine-tuned version of LLaVa trained on less data performed better, highlighting the need for validation sets during LoRA fine-tuning to prevent overtraining.


To mitigate the prompt dependency of ANLS and better assess the true impact of fine-tuning the LVLM, we developed a prompt-independent version of DMDBench in which each query elicits a single measurement or unit as the answer. Model predictions were stored in a structured CSV file and evaluated in terms of numeric, unit, and averaged word-level accuracies, with the latter being obtained by averaging the results of the numeric and unit accuracies. As shown in Table~\ref{tab:full_benchmark}, the fine-tuned InternVL model trained on 50\% of the data achieved the highest performance.

\section{Conclusion and Future Work}

This work presented CAD2DMD-SET, an automated pipeline for generating synthetic datasets aimed at improving large vision-language models~(LVLMs) performance in reading digital measurement devices~(DMDs) in visually elaborate conditions, usually present in images captured by head-mounted cameras.
By combining photorealistic rendering of CAD-based devices with contextual image composition, the tool addresses the lack of dedicated data for this visually complex and practically relevant task.
Additionally, our real-world validation set, DMDBench, provides a benchmark to assess the effectiveness of these models under realistic conditions that involve visual noise, occlusion, and motion blur, with benchmark results demonstrating substantial limitations in current LVLMs.
Furthemore, fine-tuning experiments underscore the effectiveness of the CAD2DMD-SET training dataset in enhancing the robustness and performance of LVLMs under these challenging conditions.

This study highlights certain limitations that open potential avenues for future exploration.
Enhancing the realism of the synthetic data, in particular adding material wear-and-tear and scratches to the 3D CAD models; expanding model evaluation through broader fine-tuning; better prompt engineering~\cite{gu2023survey}, and diversifying annotation types to encompass different tasks, (e.g. measurement range checking and multi-image measurement comparison), and evaluation metrics, could improve both the dataset’s applicability and the robustness of performance assessments.
Moreover, further investigation into the impact of scene composition quality on model outcomes may offer valuable insights.
Finally, OCR systems cannot extract non-alphanumeric readouts~(\textit{e.g.} signal frequency) from instruments such as oscilloscopes, however, VQA models can \emph{potentially} interpret and report them.
Leveraging CAD2DMD-SET’s extensibility, we plan to incorporate at least one such DMD into a forthcoming sample dataset.

\section*{Acknowledgment}

This work was supported by LARSyS FCT funding~(DOI: 10.54499/LA/P/0083/2020, 10.54499/UIDP/50009/2020, and 10.54499/UIDB/50009/2020)

\bibliographystyle{unsrt}  
\bibliography{main}  

\begin{thebibliography}{10}

\bibitem{jiang2024effectiveness}
Yao Jiang, Xinyu Yan, Ge-Peng Ji, Keren Fu, Meijun Sun, Huan Xiong, Deng-Ping Fan, and Fahad~Shahbaz Khan.
\newblock Effectiveness assessment of recent large vision-language models.
\newblock {\em Visual Intelligence}, 2024.

\bibitem{vo2025vision}
An~Vo, Khai-Nguyen Nguyen, Mohammad~Reza Taesiri, Vy~Tuong Dang, Anh~Totti Nguyen, and Daeyoung Kim.
\newblock Vision language models are biased.
\newblock {\em arXiv preprint arXiv:2505.23941}, 2025.

\bibitem{my_paper}
Álvaro Patrício, João Valente, Atabak Dehban, Inês Cadilha, Daniel Reis, and Rodrigo Ventura.
\newblock Ai-powered augmented reality for satellite assembly, integration and test.
\newblock In {\em IEEE International Conference on Artificial Intelligence and eXtended and Virtual Reality (AIxVR)}, 2025.

\bibitem{guassian_paper}
Nataliya Shakhovska, Bohdan Sydor, Solomiia Liaskovska, Olga Duran, Yevgen Martyn, and Volodymyr Vira.
\newblock High-fidelity synthetic data generation framework for unique objects detection.
\newblock {\em Computation}, 2025.

\bibitem{Unity_Perception}
Steve Borkman, Adam Crespi, Saurav Dhakad, Sujoy Ganguly, Jonathan Hogins, You-Cyuan Jhang, Mohsen Kamalzadeh, Bowen Li, Steven Leal, Pete Parisi, Cesar Romero, Wesley Smith, Alex Thaman, Samuel Warren, and Nupur Yadav.
\newblock Unity perception: Generate synthetic data for computer vision.
\newblock {\em arXiv preprint arXiv:2107.04259}, 2021.

\bibitem{Libcom}
Bcmi.
\newblock Libcom: Image composition toolbox.
\newblock \url{https://github.com/bcmi/libcom}.
\newblock Accessed: 2025-05-24.

\bibitem{Control-Com}
Bo~Zhang, Yuxuan Duan, Jun Lan, Yan Hong, Huijia Zhu, Weiqiang Wang, and Li~Niu.
\newblock Controlcom: Controllable image composition using diffusion model.
\newblock {\em arXiv preprint arXiv:2308.10040}, 2023.

\bibitem{FOPA}
Li~Niu, Qingyang Liu, Zhenchen Liu, and Jiangtong Li.
\newblock Fast object placement assessment.
\newblock {\em arXiv preprint arXiv:2205.14280}, 2022.

\bibitem{OPA}
Liu Liu, Zhenchen Liu, Bo~Zhang, Jiangtong Li, Li~Niu, Qingyang Liu, and Liqing Zhang.
\newblock Opa: Object placement assessment dataset.
\newblock {\em arXiv preprint arXiv:2107.01889}, 2022.

\bibitem{VLMEvalkit}
Haodong Duan, Junming Yang, Yuxuan Qiao, Xinyu Fang, Lin Chen, Yuan Liu, Xiaoyi Dong, Yuhang Zang, Pan Zhang, Jiaqi Wang, et~al.
\newblock Vlmevalkit: An open-source toolkit for evaluating large multi-modality models.
\newblock In {\em Proceedings of the 32nd ACM international conference on multimedia}, 2024.

\bibitem{EasyOCRGit}
JaidedAI.
\newblock Easyocr.
\newblock \url{https://github.com/JaidedAI/EasyOCR}.
\newblock Accessed: 2025-06-24.

\bibitem{otsu1979threshold}
Nobuyuki Otsu.
\newblock A threshold selection method from gray-level histograms.
\newblock {\em IEEE Transactions on Systems, Man, and Cybernetics}, 1979.

\bibitem{DSEG}
Keshikan.
\newblock Dseg: Font series for seven-segment display.
\newblock \url{https://github.com/keshikan/DSEG}, 2019.
\newblock Accessed: 2025-06-19.

\bibitem{indoor_mit}
Ariadna Quattoni and Antonio Torralba.
\newblock Recognizing indoor scenes.
\newblock In {\em IEEE Conference on Computer Vision and Pattern Recognition (CVPR)}, 2009.

\bibitem{ade20k}
Bolei Zhou, Hang Zhao, Xavier Puig, Tete Xiao, Sanja Fidler, Adela Barriuso, and Antonio Torralba.
\newblock Semantic understanding of scenes through the ade20k dataset.
\newblock {\em International Journal of Computer Vision}, 2019.

\bibitem{hu2023image}
Fengling Hu, Andrew~A Chen, Hannah Horng, Vishnu Bashyam, Christos Davatzikos, Aaron Alexander-Bloch, Mingyao Li, Haochang Shou, Theodore~D Satterthwaite, Meichen Yu, et~al.
\newblock Image harmonization: A review of statistical and deep learning methods for removing batch effects and evaluation metrics for effective harmonization.
\newblock {\em NeuroImage}, 274, 2023.

\bibitem{Pixtral12B}
Pravesh Agrawal, Szymon Antoniak, Emma~Bou Hanna, Baptiste Bout, Devendra Chaplot, Jessica Chudnovsky, Diogo Costa, Baudouin De~Monicault, Saurabh Garg, Theophile Gervet, et~al.
\newblock Pixtral 12b.
\newblock {\em arXiv preprint arXiv:2410.07073}, 2024.

\bibitem{Llava}
Haotian Liu, Chunyuan Li, Yuheng Li, and Yong~Jae Lee.
\newblock Improved baselines with visual instruction tuning.
\newblock In {\em IEEE Conference on Computer Vision and Pattern Recognition (CVPR)}, 2024.

\bibitem{InternVL}
Zhe Chen, Weiyun Wang, Yue Cao, Yangzhou Liu, Zhangwei Gao, Erfei Cui, Jinguo Zhu, Shenglong Ye, Hao Tian, Zhaoyang Liu, Lixin Gu, Xuehui Wang, et~al.
\newblock Expanding performance boundaries of open-source multimodal models with model, data, and test-time scaling.
\newblock {\em arXiv preprint arXiv:2412.05271}, 2025.

\bibitem{lvlm_training_paradigm}
Xiaorui Ma, Haoran Xie, and S.~{Joe Qin}.
\newblock Efficiently integrate large language models with visual perception: A survey from the training paradigm perspective.
\newblock {\em Information Fusion}, 2025.

\bibitem{LoRA}
Edward~J Hu, yelong shen, Phillip Wallis, Zeyuan Allen-Zhu, Yuanzhi Li, Shean Wang, Lu~Wang, and Weizhu Chen.
\newblock Lo{RA}: Low-rank adaptation of large language models.
\newblock In {\em International Conference on Learning Representations (ICLR)}, 2022.

\bibitem{anls_metric}
David Peer, Philemon Schöpf, Volckmar Nebendahl, Alexander Rietzler, and Sebastian Stabinger.
\newblock Anls* -- a universal document processing metric for generative large language models.
\newblock {\em arXiv preprint arXiv:2402.03848}, 2025.

\bibitem{TextVQA}
Amanpreet Singh, Vivek Natarajan, Meet Shah, Yu~Jiang, Xinlei Chen, Dhruv Batra, Devi Parikh, and Marcus Rohrbach.
\newblock Towards vqa models that can read.
\newblock In {\em IEEE Conference on Computer Vision and Pattern Recognition (CVPR)}, 2019.

\bibitem{COCO}
Tsung-Yi Lin, Michael Maire, Serge Belongie, James Hays, Pietro Perona, Deva Ramanan, Piotr Doll{\'a}r, and C~Lawrence Zitnick.
\newblock Microsoft coco: Common objects in context.
\newblock In {\em European Conference on Computer Vision (ECCV)}, 2014.

\bibitem{GQA}
Drew~A Hudson and Christopher~D Manning.
\newblock Gqa: A new dataset for real-world visual reasoning and compositional question answering.
\newblock In {\em IEEE Conference on Computer Vision and Pattern Recognition (CVPR)}, 2019.

\bibitem{OCR-VQA}
Anand Mishra, Shashank Shekhar, Ajeet~Kumar Singh, and Anirban Chakraborty.
\newblock Ocr-vqa: Visual question answering by reading text in images.
\newblock In {\em International Conference on Document Analysis and Recognition (ICDAR)}, 2019.

\bibitem{VG}
Ranjay Krishna, Yuke Zhu, Oliver Groth, Justin Johnson, Kenji Hata, Joshua Kravitz, Stephanie Chen, Yannis Kalantidis, Li-Jia Li, David~A Shamma, et~al.
\newblock Visual genome: Connecting language and vision using crowdsourced dense image annotations.
\newblock {\em International journal of computer vision}, 2017.

\bibitem{ai2d}
Aniruddha Kembhavi, Mike Salvato, Eric Kolve, Minjoon Seo, Hannaneh Hajishirzi, and Ali Farhadi.
\newblock A diagram is worth a dozen images.
\newblock In {\em European Conference on Computer Vision (ECCV)}. Springer, 2016.

\bibitem{chartqa}
Ahmed Masry, Xuan~Long Do, Jia~Qing Tan, Shafiq Joty, and Enamul Hoque.
\newblock Chartqa: A benchmark for question answering about charts with visual and logical reasoning.
\newblock In {\em Findings of the Association for Computational Linguistics}, 2022.

\bibitem{docvqa}
Minesh Mathew, Dimosthenis Karatzas, and CV~Jawahar.
\newblock Docvqa: A dataset for vqa on document images.
\newblock In {\em IEEE Winter Conference on Applications of Computer Vision (WACV)}, 2021.

\bibitem{dvqa}
Kushal Kafle, Brian Price, Scott Cohen, and Christopher Kanan.
\newblock Dvqa: Understanding data visualizations via question answering.
\newblock In {\em IEEE Conference on Computer Vision and Pattern Recognition (CVPR)}, 2018.

\bibitem{sam}
Alexander Kirillov, Eric Mintun, Nikhila Ravi, Hanzi Mao, Chloe Rolland, Laura Gustafson, Tete Xiao, Spencer Whitehead, Alexander~C Berg, Wan-Yen Lo, et~al.
\newblock Segment anything.
\newblock In {\em IEEE International Conference on Computer Vision (ICCV)}, 2023.

\bibitem{donut}
Geewook Kim, Teakgyu Hong, Moonbin Yim, JeongYeon Nam, Jinyoung Park, Jinyeong Yim, Wonseok Hwang, Sangdoo Yun, Dongyoon Han, and Seunghyun Park.
\newblock Ocr-free document understanding transformer.
\newblock In {\em European Conference on Computer Vision (ECCV)}. Springer, 2022.

\bibitem{geoqa}
Jiaqi Chen, Jianheng Tang, Jinghui Qin, Xiaodan Liang, Lingbo Liu, Eric Xing, and Liang Lin.
\newblock Geoqa: A geometric question answering benchmark towards multimodal numerical reasoning.
\newblock In {\em Findings of the Association for Computational Linguistics}, 2021.

\bibitem{stability}
Wenjuan Han, Bo~Pang, and Ying~Nian Wu.
\newblock Robust transfer learning with pretrained language models through adapters.
\newblock In {\em Proceedings of the 59th Annual Meeting of the Association for Computational Linguistics and the 11th International Joint Conference on Natural Language Processing}, 2021.

\bibitem{gu2023survey}
Jindong Gu, Zhen Han, Shuo Chen, Ahmad Beirami, Bailan He, Gengyuan Zhang, Ruotong Liao, Yao Qin, Volker Tresp, and Philip Torr.
\newblock A systematic survey of prompt engineering on vision-language foundation models.
\newblock {\em arXiv preprint arXiv:2307.12980}, 2023.

\end{thebibliography}

\newpage

\appendix

\section{Supplementary Material}

\subsection{Real World Conditions Emulation}

\begin{figure}[H]
    \centering
    \includegraphics[width=0.7\linewidth]{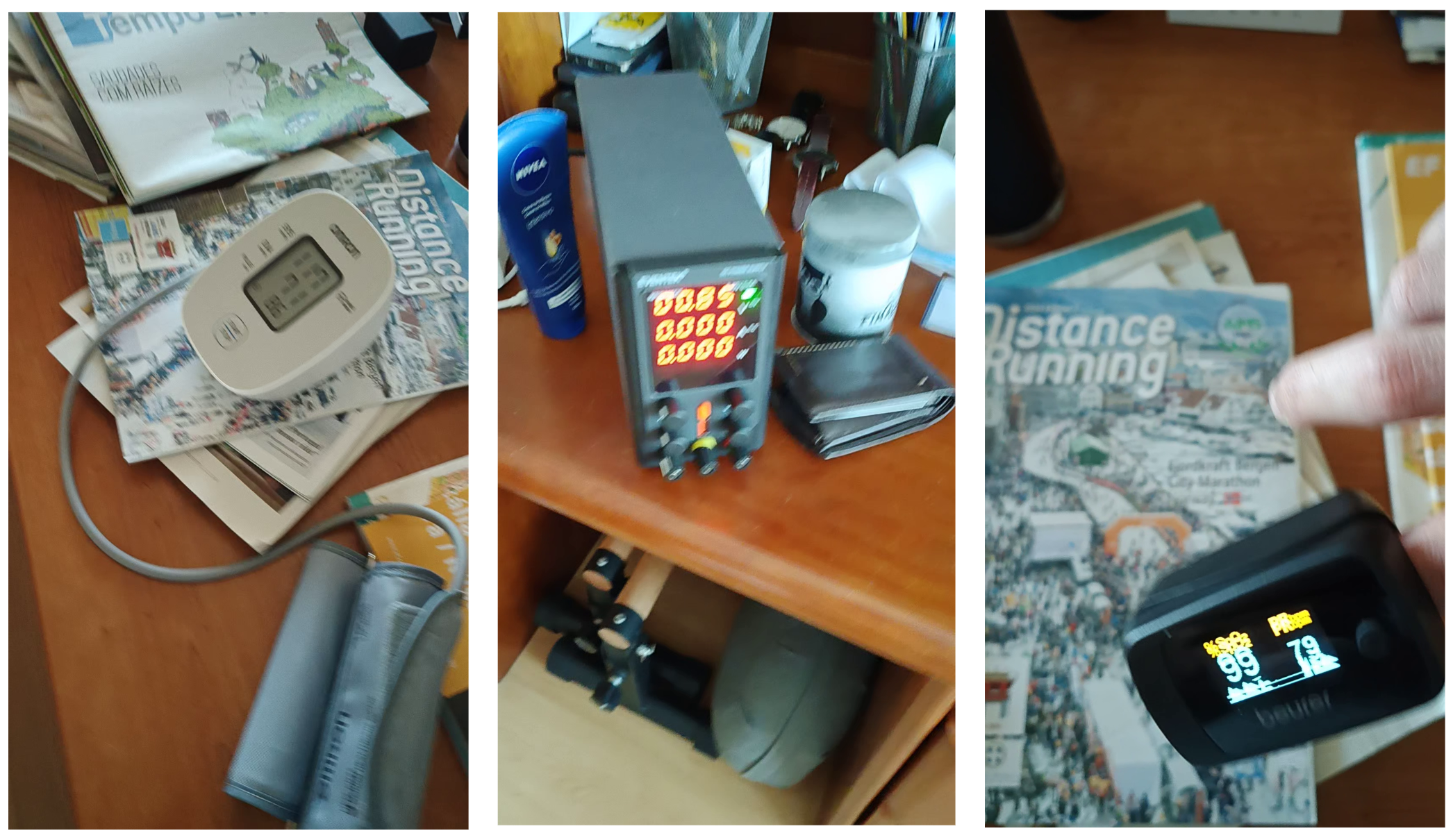}
    \caption{Challenging Motion Blur DMDBench Examples}
    \label{fig:motion_blur}
\end{figure}

\begin{figure}[H]
    \centering
    \includegraphics[width=0.7\linewidth]{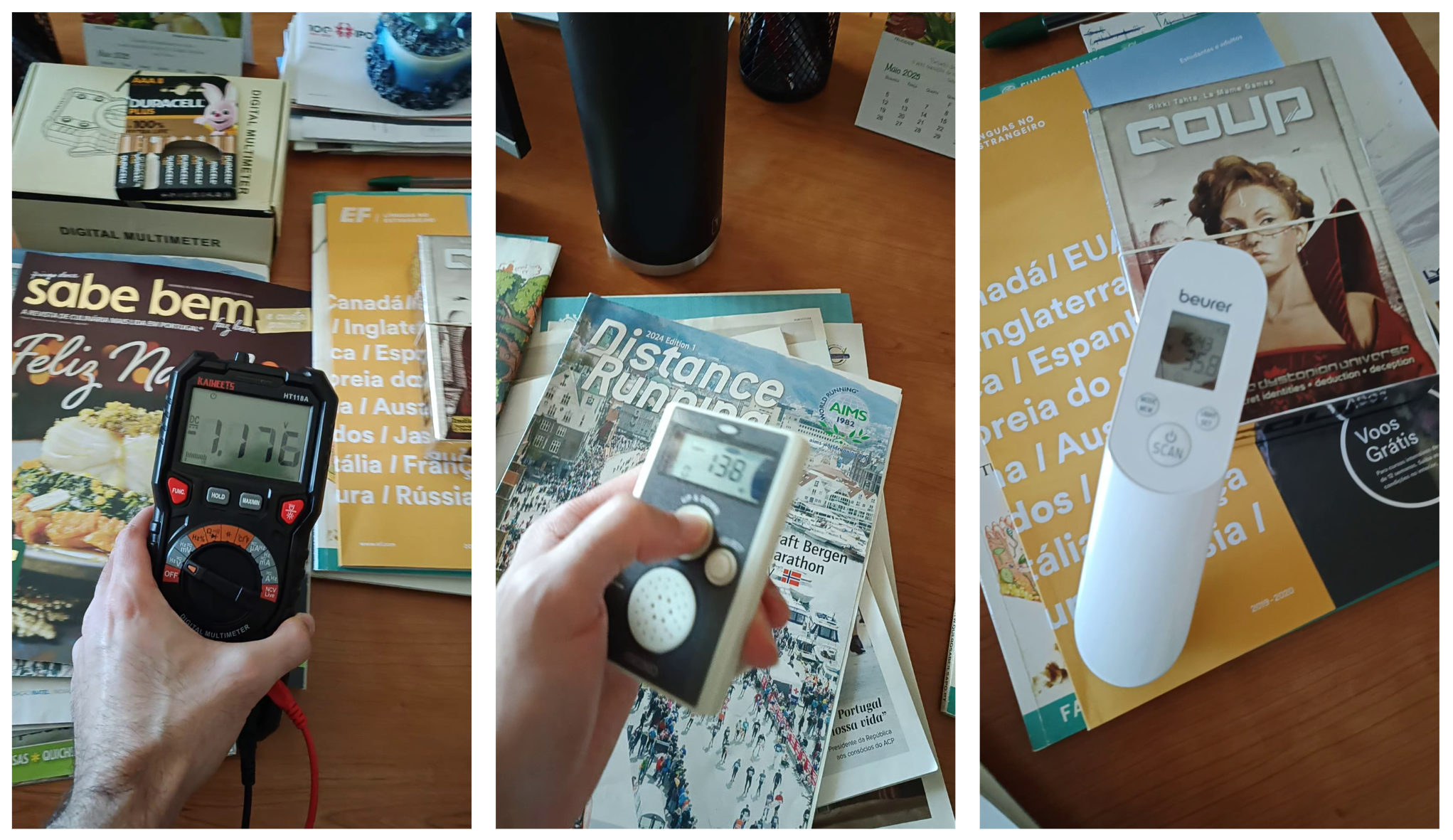}
    \caption{Challenging Lighting DMDBench Examples}
    \label{fig:lighting}
\end{figure}

\begin{figure}[H]
    \centering
    \includegraphics[width=0.7\linewidth]{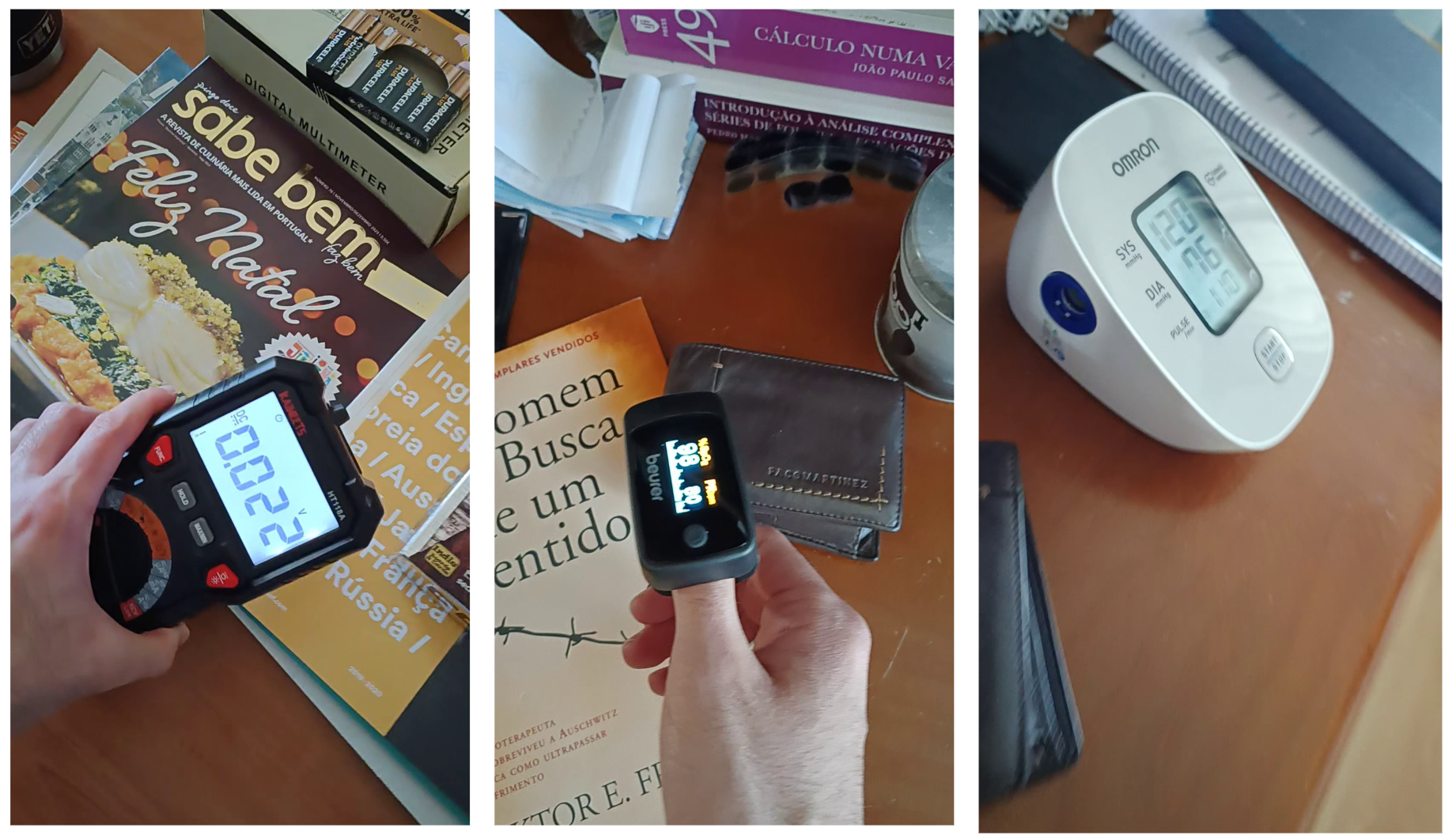}
    \caption{Challenging Orientation DMDBench Examples}
    \label{fig:orientation}
\end{figure}

\begin{figure}[H]
    \centering
    \includegraphics[width=0.7\linewidth]{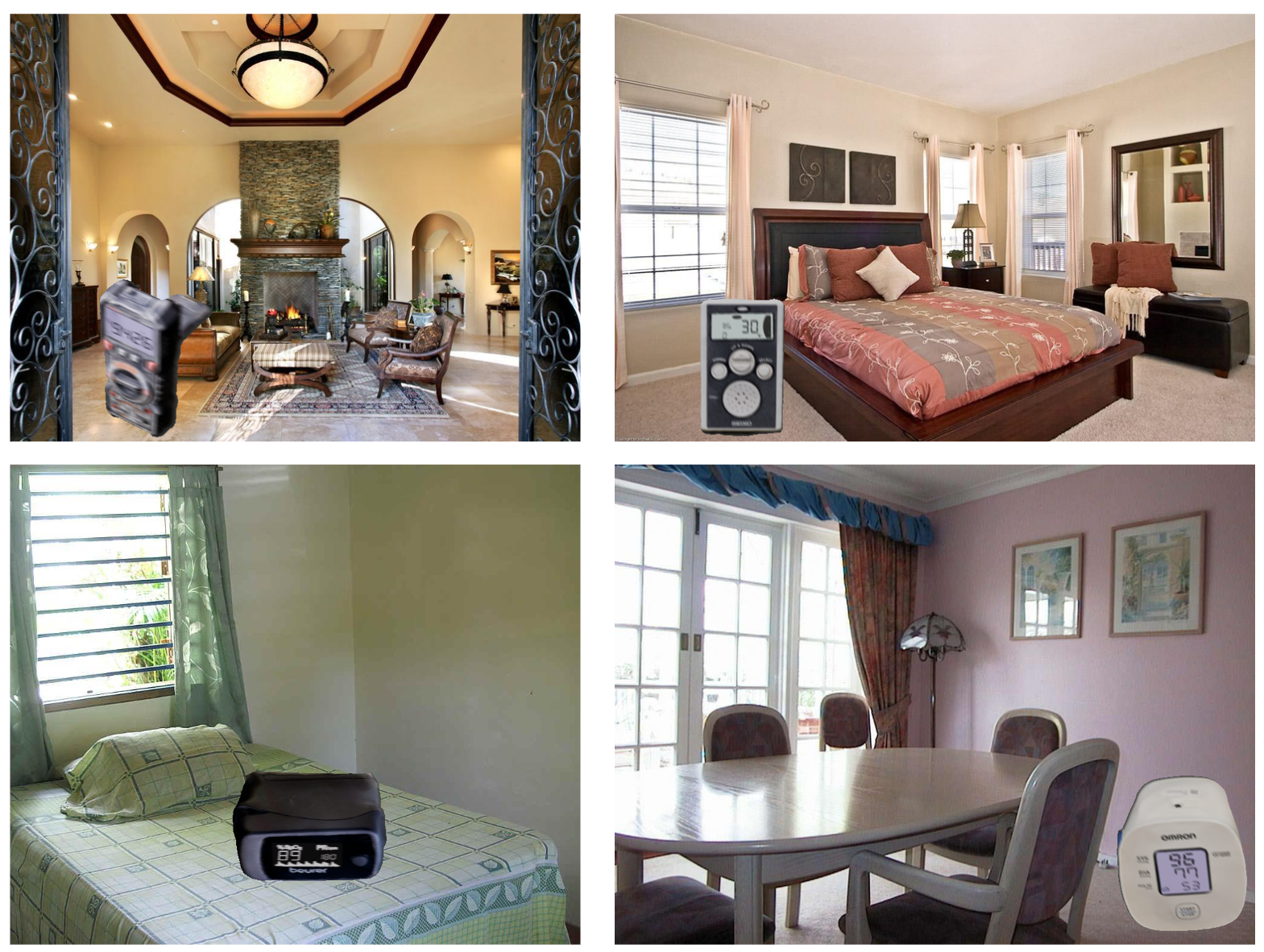}
    \caption{Training Dataset Examples}
    \label{fig:training}
\end{figure}

The DMDBench was created with the intention of emulating realistic and imperfect conditions, incorporating variations in scale, orientation, lighting, background clutter, and motion blur caused by user movements, commonly encountered in head-mounted AR devices. 
To better illustrate how DMDBench evaluates large vision–language models under such challenging conditions, examples are shown in Fig.~\ref{fig:motion_blur},~\ref{fig:lighting}, and~\ref{fig:orientation}, together with the corresponding challenges they are designed to replicate.
Furthermore, examples of training images generated by the CAD2DMD-SET tool, designed to simulate such conditions, are presented in Fig.~\ref{fig:training}.

\vspace{-1.5cm}

\vspace{1.5cm}

\vspace{-3cm}

\subsection{Training Improvements and Limitations}

\begin{figure}[H]
\centering
\caption{DMDBench Improvement Examples for InternVL2.5-26B. \textcolor{Green}{Green text} is correct answer, \textcolor{red}{red text} shows wrong answer.}
\label{fig:improvement_internvl}
\begin{tabular}{@{}p{0.48\textwidth}@{\hskip 0.04\textwidth}p{0.48\textwidth}@{}}

\begin{minipage}[t]{\linewidth}
\centering
\includegraphics[width=0.6\linewidth]{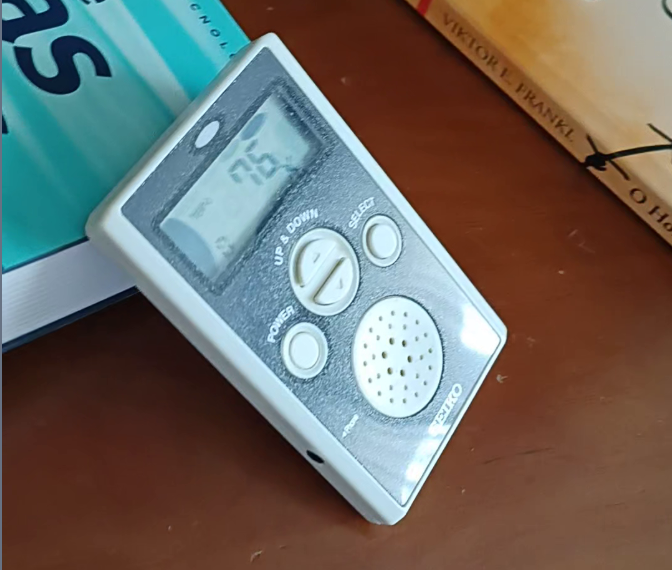}
\vspace{0.5em}

\smaller{\textbf{(a)Question:} What is the value for TEMPO? Answer with a single word, using only the measurement or unit, depending on the question.}
\end{minipage}
&
\begin{minipage}[t]{\linewidth}
\centering
\includegraphics[width=0.6\linewidth]{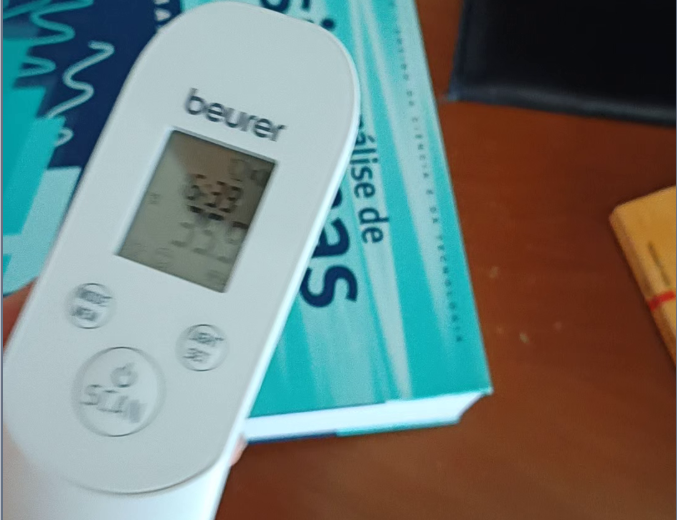}
\vspace{0.5em}

\smaller{\textbf{(b)Question:} What is the value for Temperature? Answer with a single word, using only the measurement or unit, depending on the question.}
\end{minipage}
\\
\midrule

\multicolumn{2}{l}{\textbf{Challenging Condition:}} \\
\makecell[c]{\smaller{Side orientation and intense light setting}} & \makecell[c]{ \smaller{Screen reflections}} \\
\midrule

\multicolumn{2}{l}{\textbf{Ground-Truth Answer:}} \\
\makecell[c]{\smaller{76}} & \makecell[c]{\smaller{35.9}}\\
\midrule

\multicolumn{2}{l}{\textbf{Non-fine-tuned Model Answer:}} \\ 
\makecell[c]{\smaller{\textcolor{red}{75}}} & \makecell[c]{\smaller{\textcolor{red}{36.9}}} \\
\midrule

\multicolumn{2}{l}{\textbf{Fine-tuned Model Answer:}} \\
\makecell[c]{\smaller{\textcolor{Green}{76}}} & \makecell[c]{\smaller{\textcolor{Green}{35.9}}} \\
\end{tabular}
\end{figure}

\begin{figure}[H]
\centering
\caption{DMDBench Improvement Examples for LLaVA-v1.5-13B. \textcolor{Green}{Green text} is correct answer, \textcolor{red}{red text} shows wrong answer.}
\label{fig:improvement_llava}
\begin{tabular}{@{}p{0.48\textwidth}@{\hskip 0.04\textwidth}p{0.48\textwidth}@{}}

\begin{minipage}[t]{\linewidth}
\centering
\includegraphics[width=0.6\linewidth]{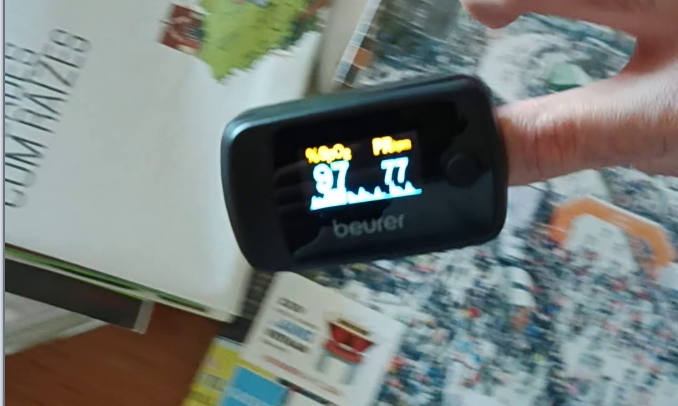}
\vspace{0.5em}

\smaller{\textbf{(a)Question:} What unit is used for Pulse Rate? Answer with a single word, using only the measurement or unit, depending on the question.}
\end{minipage}
&
\begin{minipage}[t]{\linewidth}
\centering
\includegraphics[width=0.6\linewidth]{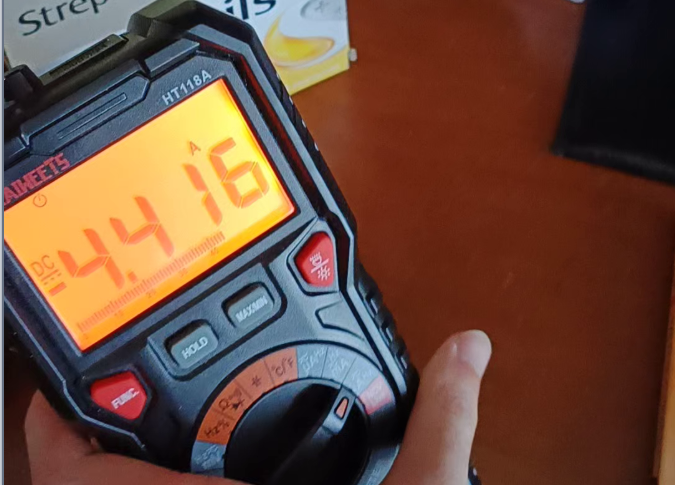}
\vspace{0.5em}

\smaller{\textbf{(b)Question:} What is the value for DC Current? Answer with a single word, using only the measurement or unit, depending on the question.}
\end{minipage}
\\
\midrule

\multicolumn{2}{l}{\textbf{Challenging Condition:}} \\
\makecell[c]{\smaller{Small scale and slight motion blur}} & \makecell[c]{\smaller{Side orientation and slight motion blur}} \\
\midrule

\multicolumn{2}{l}{\textbf{Ground-Truth Answer:}} \\
\makecell[c]{\smaller{BPM}} & \makecell[c]{\smaller{-4.416}}\\
\midrule

\multicolumn{2}{l}{\textbf{Non-fine-tuned Model Answer:}} \\ 
\makecell[c]{\smaller{\textcolor{red}{1/min}}} & \makecell[c]{\smaller{\textcolor{red}{4.16}}} \\
\midrule

\multicolumn{2}{l}{\textbf{Fine-tuned Model Answer:}} \\
\makecell[c]{\smaller{\textcolor{red}{202} \textcolor{Green}{BPM}}} & \makecell[c]{\smaller{\textcolor{red}{-}\textcolor{Green}{4.416} \textcolor{red}{A}}} \\
\end{tabular}
\end{figure}

\begin{figure}[H]
\centering
\caption{DMDBench Challenging Examples for InternVL2.5-26B. \textcolor{Green}{Green text} is correct answer, \textcolor{red}{red text} shows wrong answer.}
\label{fig:limitations_internvl}
\begin{tabular}{@{}p{0.48\textwidth}@{\hskip 0.04\textwidth}p{0.48\textwidth}@{}}

\begin{minipage}[t]{\linewidth}
\centering
\includegraphics[width=0.6\linewidth]{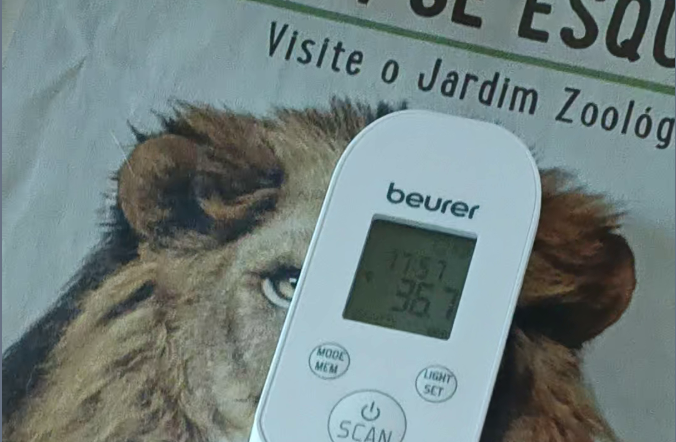}
\vspace{0.5em}

\smaller{\textbf{(a)Question:} What is the value for Time? Answer with a single word, using only the measurement or unit, depending on the question.}
\end{minipage}
&
\begin{minipage}[t]{\linewidth}
\centering
\includegraphics[width=0.6\linewidth]{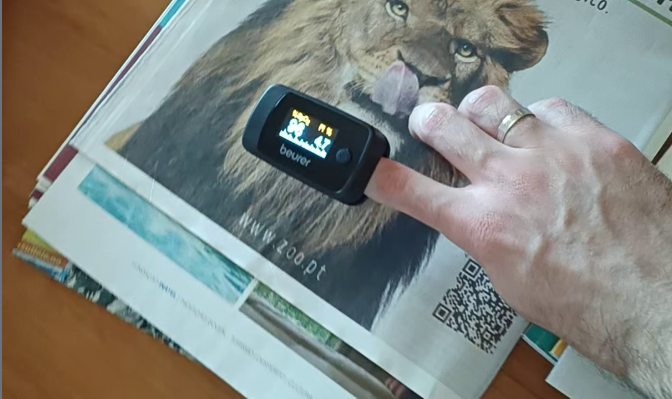}
\vspace{0.5em}

\smaller{\textbf{(b)Question:} What is the value for SpO2? Answer with a single word, using only the measurement or unit, depending on the question.}
\end{minipage}
\\
\midrule

\multicolumn{2}{l}{\textbf{Challenging Condition:}} \\
\makecell[c]{\smaller{Low-light setting}} & \makecell[c]{\smaller{Small scale and side orientation}}\\
\midrule

\multicolumn{2}{l}{\textbf{Ground-Truth Answer:}} \\
\makecell[c]{\smaller{17:57}} & \makecell[c]{\smaller{96}}\\
\midrule

\multicolumn{2}{l}{\textbf{Non-fine-tuned Model Answer:}} \\ 
\makecell[c]{\smaller{\textcolor{red}{11:57}}} & \makecell[c]{\smaller{\textcolor{red}{98}}} \\
\midrule

\multicolumn{2}{l}{\textbf{Fine-tuned Model Answer:}} \\
\makecell[c]{\smaller{\textcolor{red}{11:57}}} & \makecell[c]{\smaller{\textcolor{red}{99}}} \\
\end{tabular}
\end{figure}

\begin{figure}[H]
\centering
\caption{DMDBench Challenging Examples for LLaVA-v1.5-13B. \textcolor{Green}{Green text} is correct answer, \textcolor{red}{red text} shows wrong answer.}
\label{fig:limitations_llava}
\begin{tabular}{@{}p{0.48\textwidth}@{\hskip 0.04\textwidth}p{0.48\textwidth}@{}}

\begin{minipage}[t]{\linewidth}
\centering
\includegraphics[width=0.6\linewidth]{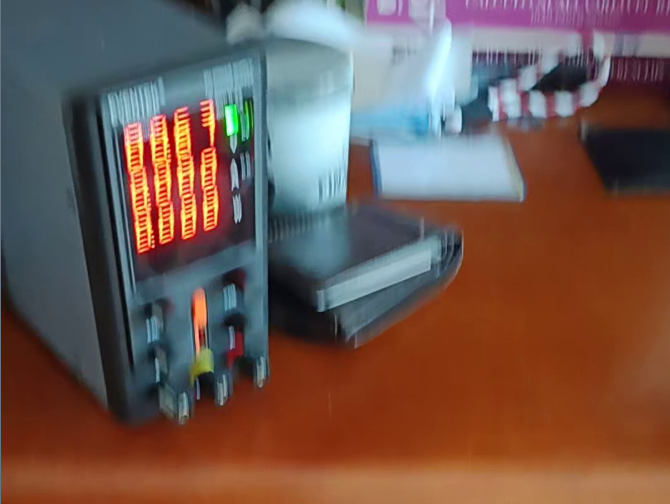}
\vspace{0.5em}

\smaller{\textbf{(a)Question:} What is the value for Voltage? Answer with a single word, using only the measurement or unit, depending on the question.}
\end{minipage}
&
\begin{minipage}[t]{\linewidth}
\centering
\includegraphics[width=0.6\linewidth]{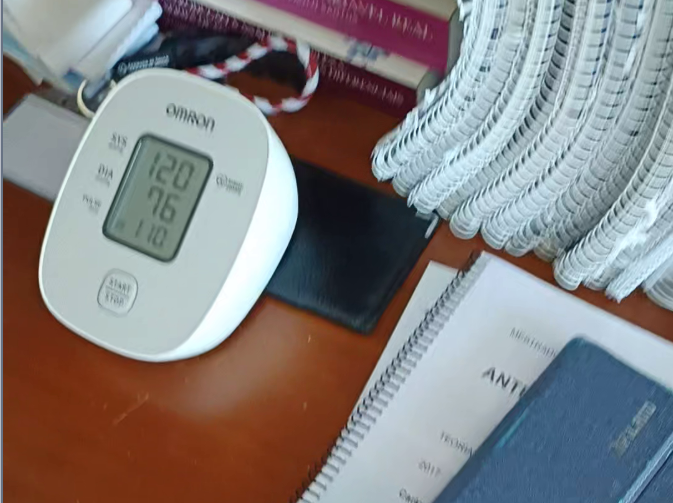}
\vspace{0.5em}

\smaller{\textbf{(b)Question:} What unit is used for Diastolic Blood Pressure? Answer with a single word, using only the measurement or unit, depending on the question.}
\end{minipage}
\\
\midrule

\multicolumn{2}{l}{\textbf{Challenging Condition:}} \\
\makecell[c]{\smaller{Motion Blur}} & \makecell[c]{\smaller{Motion blur and side orientation}}\\
\midrule

\multicolumn{2}{l}{\textbf{Ground-Truth Answer:}} \\
\makecell[c]{\smaller{00.67}} & \makecell[c]{\smaller{mmHg}}\\
\midrule

\multicolumn{2}{l}{\textbf{Non-fine-tuned Model Answer:}} \\ 
\makecell[c]{\smaller{\textcolor{red}{120}}} & \makecell[c]{\smaller{\textcolor{red}{100}}} \\
\midrule

\multicolumn{2}{l}{\textbf{Fine-tuned Model Answer:}} \\
\makecell[c]{\smaller{\textcolor{red}{08.31}}} & \makecell[c]{\smaller{\textcolor{red}{76}}} \\
\end{tabular}
\end{figure}

Fine-tuning experiments demonstrated the effectiveness of the CAD2DMD-SET training dataset in improving the robustness and performance of LVLMs under challenging real-world conditions. Nonetheless, certain limitations persist, particularly in low-light environments and scenarios with severe motion blur. To illustrate both the achieved improvements and some remaining failure cases, example images are presented in Fig.~\ref{fig:improvement_internvl},~\ref{fig:improvement_llava},~\ref{fig:limitations_internvl} and~\ref{fig:limitations_llava}.

\subsection{Ablation Studies}

In order to assess the influence of specific CAD2DMD-SET features on the performance of large vision–language models trained with its dataset, four ablation studies were conducted, focusing on object placement and scale, motion blur, one-word label training, and inter-device generalization, respectively.

\subsubsection{Object Placement and Scale}

In CAD2DMD-SET's image composer, the FOPA model is used for contextually appropriate placement and scale. 
However, due to resizing and reshaping boundary box operations, needed to achieve readable foreground objects, the realism of the final dataset is affected. 
As such, in order to evaluate if the models benefit from a realistic placement, given by FOPA, an ablation study was conducted, in which InternVL2.5-26B was evaluated on DMDBench, under two different scenarios: fine-tuned on 10,000 images from the original dataset using FOPA, and fine-tuned on 10,000 images where the foreground objects were randomly placed and scaled. The ablation study results can be visualized in Table~\ref{tab:fopa_ablation}.

\begin{table}[H]
\caption{InternVL2.5-26B Object Placement and Scale Ablation Study Results}
\small
\label{tab:fopa_ablation}
\centering
\begin{tabular}{|c|c|c|}
\hline
\textbf{Fine-tuning Dataset}   & \textbf{Single Image Generation Time (s)} & \textbf{ANLS Score (\%)}  \\ \hline
\textbf{None} &  -            & 32.92             \\ \hline
\textbf{FOPA}  & 1.4              &  \textbf{95.16}            \\ \hline
\textbf{Random}  & \textbf{0.2}               & 78.17             \\ \hline
\end{tabular}
\end{table}

Both methods improved performance relative to the non-fine-tuned model. 
In particular, fine-tuning InternVL2.5-26B on the FOPA dataset increased the ANLS score from 78.17\% to 95.16\%, indicating that the model benefits more from realistic object placement rather than solely from exposure to rendered objects against varied backgrounds.
Although fine-tuning with the FOPA dataset yields superior results, generating a single image using this dataset is approximately seven times longer than when employing random positioning and scaling. The tool provides both options, enabling users to select their preferred image composition method according to their priority, whether optimizing model performance or minimizing dataset generation time.




\subsubsection{Motion Blur}

Motion blur is a common and challenging real-world condition, particularly in head-mounted cameras and augmented reality (AR) applications.
To assess the impact of motion blur in the training set on model performance, an ablation study was conducted in which InternVL2.5-26B was evaluated on DMDBench under two scenarios: fine-tuned on 10,000 images from the original dataset, where approximately 20\% of the images exhibit motion blur of varying intensities and directions, and fine-tuned on 10,000 images without motion blur. 
The ablation study results are presented in Table~\ref{tab:mb_ablation}.

\begin{table}[H]
\caption{InternVL2.5-26B Motion Blur Ablation Study Results}
\small
\label{tab:mb_ablation}
\centering
\begin{tabular}{|c|c|c|}
\hline
\textbf{Fine-tuning Dataset} & \textbf{Motion Blur Presence (\%)}   & \textbf{ANLS Score (\%)}  \\ \hline
\textbf{None} & -             & 32.92             \\ \hline
\textbf{Original} & 20               &  \textbf{95.16}            \\ \hline
\textbf{No Motion Blur} & 0                & 88.89             \\ \hline
\end{tabular}
\end{table}

Both methods achieved performance gains over the non-fine-tuned baseline. While fine-tuning InternVL2.5-26B using a dataset without motion blur in any training images still increased the ANLS score from 32.92\% to 88.89\%, the highest performance was obtained when 20\% of the training images contained motion blur. This result highlights the positive impact of incorporating motion blur during fine-tuning for enhancing model robustness under challenging real-world conditions.

\subsubsection{One-word Label Format}

Since ANLS is highly sensitive to prompt variations, potentially masking the actual effect of fine-tuning the VLM, we developed an alternative evaluation focusing on the model’s ability to recognize measurements (numbers) and units in a less prompt-dependent manner. In this second version of DMDBench, each prompted question is designed to produce a single measurement or unit as the response and word-level, numeric, and unit accuracies are computed.
To assess the impact of label format during the fine-tuning stage of LVLMs, an ablation study was conducted in which InternVL2.5-26B was evaluated on this modified version of DMDBench under two scenarios: fine-tuned on 10,000 images from the original dataset, with the full answer label format, and fine-tuned on 10,000 images with the new one-word label format, comprising of either a measurement or unit. The ablation study results are presented in Table~\ref{tab:label_ablation}.

\begin{table}[H]
\caption{InternVL2.5-26B One-word Label Ablation Study}
\small
\label{tab:label_ablation}
\centering
\begin{tabular}{|c|c|c|c|}
\hline
\textbf{Label Format}   & \makecell{\textbf{Word-level} \\ \textbf{Accuracy}}  & \makecell{\textbf{Unit} \\ \textbf{Accuracy}} & \makecell{\textbf{Numeric} \\ \textbf{Accuracy}}\\ \hline
\textbf{None}  & 65 &  86.2 &  50.8       \\ \hline
\textbf{Full asnwer}  & 69.4 &  84.5 & 61.4 \\ \hline
\textbf{One-word}  & \textbf{77.1} & \textbf{99.6} & \textbf{62.9} \\ \hline
\end{tabular}
\end{table}

As expected, both methods outperformed the non-fine-tuned model. Notably, the model fine-tuned on the one-word label format dataset achieved an approximate 8\% increase in overall word-level accuracy, with the majority of this gain attributable to a 15\% improvement in unit-level accuracy. These results suggest that, from an OCR capability standpoint, the model benefits more from the one-word label format, which emphasizes the actual readings, than from the full-sentence label format, where sentence structure also influences performance.

\subsubsection{Inter-device Generalization}

In order to assess the model's ability to generalize when fine-tuned on our synthetic dataset, an ablation study in which InternVL2.5-26B was evaluated on a test set containing only DMDBench's power supply images, under three different scenarios: without fine-tuning, fine-tuned on 10,000 images from the original dataset comprising all six devices, and fine-tuned on 10,000 images excluding any power supply samples. 
The ablation study results are presented in Table~\ref{tab:dg_ablation}.

\begin{table}[H]
\caption{InternVL2.5-26B Inter-device Generalization Ablation Study}
\small
\label{tab:dg_ablation}
\centering
\begin{tabular}{|c|c|}
\hline
\textbf{Fine-tuning Dataset}  & \textbf{ANLS Score (\%)} \\ \hline
\textbf{None}  &   41.58      \\ \hline
\textbf{Original}    & \textbf{91.45} \\ \hline
\textbf{No Power Supply}  & 72.94 \\ \hline
\end{tabular}
\end{table}

Both fine-tuning methods outperformed the non-fine-tuned baseline, as expected. Training InternVL2.5-26B without any power supply images resulted in a decrease in ANLS score from 91.45\% to 72.94\%. Notably, a substantial improvement from 41.58\% to 72.94\% was still observed when comparing the non-fine-tuned model with the one fine-tuned without power supply images, demonstrating the inter-device generalization capability of the LVLM when fine-tuned with the synthetic dataset generated by CAD2DMD-SET.

\end{document}